\documentclass[11pt]{article}
\usepackage{eamt20}
\usepackage{times}
\usepackage{url}
\usepackage{latexsym}
\usepackage[small,bf]{caption} 
\usepackage{amsfonts}
\usepackage{xcolor}
\usepackage{booktabs}
\usepackage[printonlyused,nolist]{acronym}
\usepackage{float}
\usepackage{subcaption}
\usepackage{multirow}
\definecolor{darkblue}{rgb}{0, 0, 0.5}

\newcommand\blfootnote[1]{%
  \begingroup
  \renewcommand\thefootnote{}\footnote{#1}%
  \addtocounter{footnote}{-1}%
  \endgroup
}

\title{Learning Non-Monotonic Automatic Post-Editing of Translations from Human Orderings}

\author{Ant{\'o}nio G{\'o}is\textsuperscript{\normalfont \small *}
  \smallskip \\
  Unbabel\\
  Lisbon, Portugal \smallskip \\
  {\small \tt antoniogois@gmail.com}  \And
  Kyunghyun Cho \smallskip \\
  New York University \\
  Facebook AI \\
  New York, USA \smallskip\\
  {\small \tt kyunghyun.cho@nyu.edu} 
  \And
  Andr{\'e} Martins \smallskip \\
  Unbabel \\
  Instituto de Telecomunica\c{c}\~oes\\
  Lisbon, Portugal \smallskip \\
  {\small \tt andre.martins@unbabel.com}
  }

\date{}

\begin{document}
\maketitle
\begin{abstract}
Recent research in neural machine translation has explored flexible generation orders, as an alternative to left-to-right generation. However, training non-monotonic models brings a new complication: how to search for a good ordering when there is a combinatorial explosion of orderings arriving at the same final result? Also, how do these automatic orderings compare with the actual behaviour of human translators? Current models rely on manually built biases or are left to explore all possibilities on their own. In this paper, we analyze the orderings produced by human post-editors and use them to train an automatic post-editing system. We compare the resulting system with those trained with left-to-right  and random post-editing orderings. We observe that humans tend to follow a nearly left-to-right order, but with interesting deviations, such as preferring to start by correcting punctuation or verbs. 
\end{abstract}

\section{Introduction}
 Neural\blfootnote{\textsuperscript{*}Work partly done during a research visit at New York University.} sequence generation models have been widely adopted for tasks such as \ac{MT} \cite{bahdanau2015neural,vaswani2017attention} and automatic post-editing of translations \cite{junczys2016log-ape,chatterjee2016fbk-ape,correia2019simple,lopes2019unbabel}. These models typically generate one word at a time, and rely on a factorization that imposes a left-to-right generation ordering. Recent alternatives allow for different generation orderings \cite{welleck2019non,stern2019insertion,gu2019insertion}, or even for parallel generation of multiple tokens \cite{gu2018nonautoregressive,stern2019insertion,gu2019levenshtein,Zhou2020Understanding}, which allows exploiting dependencies among non-consecutive tokens. One potential difficulty when training non-monotonic models is how to learn a good generation ordering. There are exponentially many valid orderings to generate a given sequence, and a model should prefer those that lead to accurate translations and can be efficiently learned. In previous work, to guide the search for a good ordering, oracle policies have been provided \cite{welleck2019non}, or another kind of inductive bias such as a loss function tailored to promote certain orderings \cite{stern2019insertion}. However, no supervision has been used with orderings that go beyond simple patterns, such as left-to-right, random ordering with a uniform distribution, or a balanced binary tree.  
 
 \begin{table}[t]
\centering
\begin{tabular}{l}
\toprule
\texttt{0:} Die LMS ge{\"o}ffnet \textcolor{red}{\textit{ist}} . \quad \quad \hspace{0cm} [ I:2:\textcolor{blue}{\textbf{ist}} ] \\
\texttt{1:} Die LMS \textcolor{blue}{\textbf{ist}} ge{\"o}ffnet \textcolor{red}{\textit{ist}} . \quad   [ D:4:\textcolor{red}{\textit{ist}} ]         \\
\texttt{2:} Die LMS \textcolor{blue}{\textbf{ist}} ge{\"o}ffnet . \\
\bottomrule
\end{tabular}
\caption{Example of a small post-edit from the training set. Each action is represented by three features: its type (I for insert and D for delete), its position in the sentence and the token to insert/delete. In this example, the token marked  \textcolor{red}{\textit{in red}} needs to be removed since it is incorrectly placed. The \textcolor{blue}{\textbf{blue}} token is inserted to obtain the correct \texttt{pe}.}
\label{tab:small_example}
\end{table}
 
While prior work has focused on learning generation orderings in an unsupervised manner, in this paper we ask the question of whether human generation orderings can be a useful source of supervision. One such possible source lies in the keystrokes of humans typing. It is known that edit operations performed by human translators are not arbitrary \cite{gois2019translator2vec}. But it is not known how the orderings preferred by humans look like, or how they compare to orders learned by models.

To investigate this question, we propose a model that learns generation orderings in a supervised manner from human keystrokes. 
Since a human is free to move back and forth arbitrarily while editing text, the chosen order of operations can be used as an additional learning signal.
More specifically, we do this in the context of \textbf{\ac{APE}} \cite{simard-etal-2007-post-editing}. \ac{APE} consists in improving the output of a blackbox \ac{MT} system by automatically fixing its mistakes.
The act of post-editing text can be fully specified as a sequence of delete (\texttt{DEL}) and insert (\texttt{INS}) \textbf{actions} in given positions. Furthermore, if we do not include redundant actions in a sequence, that sequence can be arbitrarily reordered while still producing the same output. For instance, in Table~\ref{tab:small_example}, we can switch the order of the two actions, as long as we rectify to delete position 3 instead of position 4.

We compare a model trained with human orderings to others trained with left-to-right and random orderings. We show that the resulting non-monotonic \ac{APE} system learned from human orderings outperforms systems learned on random orderings and performs comparably or slightly better than a system learned with left-to-right orderings. 

\section{Dataset}

\subsection{WMT data and keystrokes}\label{sec:data_subsection}

The dataset used in this paper is the keystrokes dataset introduced by \newcite{specia2017translation} in the scope of the QT21 project. This dataset consists of triplets required to train an \ac{APE} system: \textit{source} sentences (\texttt{src}), \textit{machine-translation} outputs (\texttt{mt}) and \textit{human post-edits} (\texttt{pe}). Features about the post-editing process are also provided, including the keystroke logging. In particular, we focus on the language pair English to German (\texttt{En-De}) in the \ac{IT} domain, translated with a Phrase-Based Statistical \ac{MT} system (PBSMT) -- 
this dataset has a large intersection with the data used in the WMT 2016-18 APE shared tasks \cite{chatterjee2018findings}.
This allows for comparison with systems previously submitted to the shared task by using the exact same development and test sets,
while augmenting the training set with keystroke logging information.

\begin{table}[t]
\small
\centering
\begin{tabular}{lrrrr}
\toprule
           & size   & {\tt mt=pe}  &
           min-edit &
            human-edit \\
 \midrule
 \begin{tabular}[c]{@{}l@{}}train with\\ keystrokes\end{tabular} & 16,068 & 18.2\% & 6.6 & 14.48 \\
 \midrule
full train & 23,000 & 14.6\% & 11.8        & ---          \\
dev '16    & 1,000  & 6.0\%  & 11.3        & ---   \\
\bottomrule
\end{tabular}
\caption{WMT-APE datasets: Original training set and development set from the WMT-APE shared task, and subset of the training set also found in the dataset from \newcite{specia2017translation}. \texttt{mt=pe} is the percentage of samples where the \texttt{mt} output is already correct. min-edit is the average count of actions (\texttt{DEL} ans \texttt{INS}) required to change \texttt{mt} into \texttt{pe}, computed from Levenshtein distance. human-edit is the average count of actions computed from human keystrokes. Both average action counts exclude samples with zero actions.}
\label{tab:dataset}
\end{table}

Out of 23,000 training samples provided by the WMT 2016-17 shared tasks,
16,068 are also present in the dataset from \newcite{specia2017translation}. This intersection is obtained by requiring the same triplet (\texttt{src}, \texttt{mt}, \texttt{pe}) to be present in both datasets. Since the WMT dataset comes already pre-processed, the following pre-processing is applied to the dataset containing keystrokes, to increase their intersection: using tools from Moses \cite{koehn-etal-2007-moses}, we apply \texttt{En} punctuation-normalization to the whole triplet, followed by tokenization of the corresponding language (either \texttt{En} or \texttt{De}). Additionally, we preprocess the raw keystrokes to obtain word-level \texttt{DEL} and \texttt{INS} actions (detailed in \S\ref{sec:preprocess}).

Table~\ref{tab:dataset} shows statistics from WMT's original data and training set after intersecting with the keystrokes dataset from \newcite{specia2017translation}. We denote by min-edit the average count of \texttt{DEL} and \texttt{INS} obtained from the Levenshtein distance. Average count of human actions (human-edit) is only available for the subset of the training data found in the keystrokes dataset.
Also note that keystrokes will not be required during inference, only for training. Once a model is already trained, the only input required is a (\texttt{src}, \texttt{mt}) pair in order to predict a full sequence of actions and produce the final \texttt{pe}. This allows to use the exact same development and test sets as in the shared task, without losing any samples.

\subsection{Preprocessing raw keystrokes}\label{sec:preprocess}

\begin{table*}[t]
\centering
\small
\begin{tabular}{ll}
\toprule
\texttt{src}   &\begin{tabular}[c]{@{}l@{}}When you decrease opacity , the underlying artwork becomes visible through the surface of the object , stroke ,\\  fill , or text .\end{tabular} \\  
\midrule
\texttt{mt}    & \begin{tabular}[c]{@{}l@{}}Wenn Sie die Deckkraft verringern , wird das zugrunde liegende Bildmaterial durch die Oberfläche des Objekts , \\ Kontur , Fläche oder Text angezeigt .\end{tabular} \\
\midrule
\texttt{pe}    & \begin{tabular}[c]{@{}l@{}}Wenn Sie die Deckkraft verringern , wird das darunterliegende Bildmaterial durch die Oberfläche des Objekts ,\\ \textcolor{blue}{\textbf{der}} Kontur , \textcolor{blue}{\textbf{der}} Fläche bzw. des Textes sichtbar .\end{tabular} \\
\midrule
\textit{l2r}   & \begin{tabular}[c]{@{}l@{}}D:8:zugrunde \quad D:8:liegende \quad I:8:darunterliegende \quad I:16:\textcolor{blue}{\textbf{der}} \quad I:19:\textcolor{blue}{\textbf{der}} \quad D:21:oder \quad D:21:Text \\ D:21:angezeigt \quad I:21:bzw. \quad I:22:des \quad I:23:Textes \quad I:24:sichtbar \quad STOP\end{tabular} \\
\midrule
\textit{shuff} & \begin{tabular}[c]{@{}l@{}} D:20:oder \quad I:22:bzw. \quad D:20:Text \quad I:22:des \quad I:10:darunterliegende \quad D:8:zugrunde \quad I:23:sichtbar \\ \quad I:19:\textcolor{blue}{\textbf{der}} \quad I:24:Textes \quad I:17:\textcolor{blue}{\textbf{der}} \quad D:22:angezeigt \quad D:8:liegende \quad STOP\end{tabular} \\
\midrule
\textit{h-ord} &
  \begin{tabular}[c]{@{}l@{}}I:17:\textcolor{blue}{\textbf{der}} \quad I:20:\textcolor{blue}{\textbf{der}} \quad D:22:oder \quad I:24:bzw. \quad I:25:des \quad D:22:Text \quad I:25:Textes \quad D:8:zugrunde \\ D:8:liegende \quad I:8:darunterliegende \quad D:21:angezeigt \quad I:24:sichtbar \quad STOP\end{tabular} \\
\midrule
\textit{human} & \begin{tabular}[c]{@{}l@{}}I:17:\textcolor{blue}{\textbf{der}} \quad I:20:\textcolor{blue}{\textbf{der}} \quad D:22:oder \quad I:22:bzw. \quad  I:23:des \quad D:24:Text \quad I:24:Textes \quad D:8:zugrunde \quad\\ D:8:liegende \quad I:8:darunterliegende \quad D:25:. \quad D:24:angezeigt \quad I:24:sichtbar \quad I:25:. \quad STOP\end{tabular} \\
\bottomrule
\end{tabular}
\caption{Example of a sentence and its minimum-edit actions ordered in three different ways: left-to-right (\textit{l2r}), randomly shuffled (\textit{shuff}) and following human order (\textit{h-ord}). The unfiltered human actions are also presented (\textit{human}). We can see that the human chose to first insert the two words marked \textcolor{blue}{\textbf{in blue}}, later moving back in the sentence to edit the leftmost mistakes.}
\label{table:sample}
\end{table*}

The original keystrokes logging provides character-level changes made by the human editor. Since this information is too fine-grained for our model, we preprocess the raw keystokes to obtain word-level \texttt{DEL} and \texttt{INS}. Our starting point is the sequence of strings containing the \texttt{mt} state after each keystroke. We track which word is currently being edited and store an action to summarize the change. Replacements are represented as a \texttt{DEL} followed by \texttt{INS}. Multiple words may be changed simultaneous, either by selecting and deleting a block of words or by pasting text. Block changes assume a left-to-right sequence of actions.

Table~\ref{table:sample} contains an example of a preprocessed sequence of keystrokes in the last line (\textit{human}). When applied to the \texttt{mt}, the sentence is converted to \texttt{pe}. Note that this kind of action sequences may be impossible to re-order due to redundant actions --- for a token \texttt{t} not present in \texttt{mt} nor \texttt{pe}, the actions \texttt{INS:0:t} \texttt{DEL:0:t} cannot be switched. 

We perform an additional step to eliminate redundant actions performed by human post-editors.  Editors may take paths significantly longer than the minimum edit distance. During experiments these longer paths proved harmful for the model, so we designed a way to filter all actions which are not relevant. First, an optimal action sequence that minimizes edit (Levenshtein) distance is obtained with dynamic programming. Since by definition this sequence does not contain  redundant actions, these actions can be reordered to produce the same output. This provides a chance to experiment with different orders, such as left-to-right or human order. To align the unfiltered human actions with the minimum-edit actions,  we first match human actions to machine actions that insert/delete the same token. Then, we break ties by aligning each machine action to the human action applied to the closest position in the sentence. Note that in some cases this alignment is not possible -- for instance in Table~\ref{tab:small_example}, if the editor had moved the word \textit{ge{\"o}ffnet} instead of \textit{ist}, the alignment would have failed. However, in practice this only happens in around 1\% of the samples. In such cases, we simply keep the unfiltered human order.

For reproducibility purposes, we provide the dataset containing the (\texttt{src}, \texttt{mt}, \texttt{pe}) triplets, together with the four kinds of action sequences seen in Table~\ref{table:sample}, in \url{https://github.com/antoniogois/keystrokes_ape}.

\subsection{Analysis of action sequences}

\begin{figure*}[t]
    \centering
    \begin{subfigure}{\columnwidth}
        \includegraphics[width=\columnwidth]{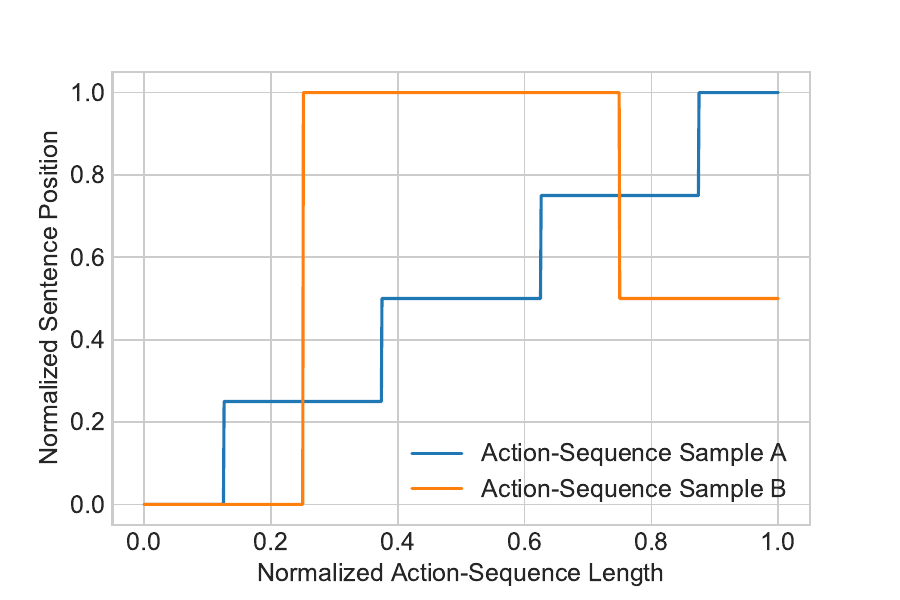}
        \label{fig:sample_orders}
    \end{subfigure}
    \begin{subfigure}{\columnwidth}
        \includegraphics[width=\columnwidth]{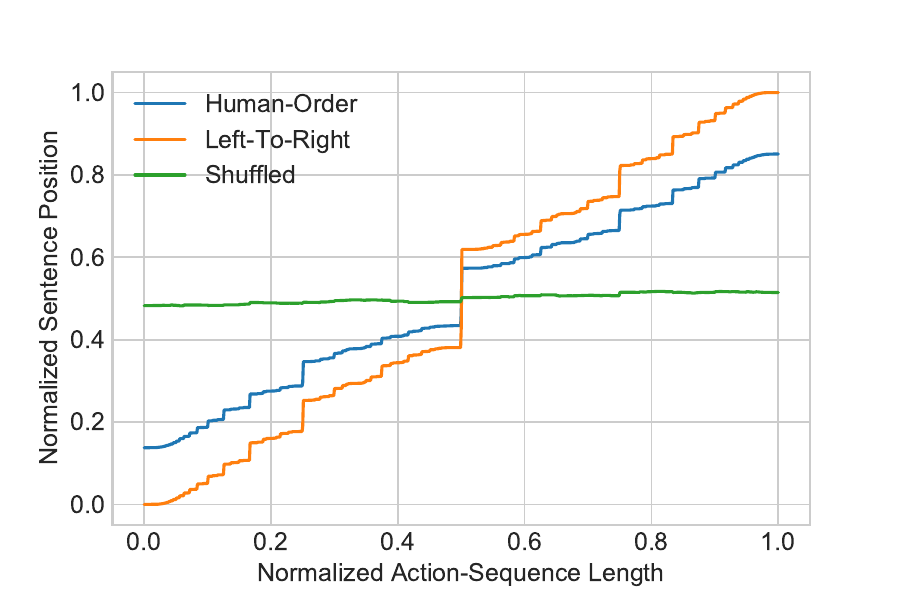}
        \label{fig:avg_orders}
    \end{subfigure}
    \caption{Relative positions of actions in a sequence. Each action-sequence length is normalized to 1, and we show the order by which actions were applied. For instance, sample A in the left-hand plot contains 3 actions, applied in a non-monotonic order --- first the leftmost action, then the rightmost and finally the middle-action (e.g. corresponding to actions in positions 3, 8 and 5 of a sentence). Samples A and B represent respectively 2.0\% and 4.2\% of the human action-sequences. The right-hand plot shows the average of all sequences, in human order, left-to-right and shuffled. Human-order tends to follow a left-to-right order, but not as strict as the artificial left-to-right. The shuffled sequences do not follow any pattern, as expected.}
    \label{fig:orders_plots}
\end{figure*}

Given the actions provided by the minimum-edit distance between \texttt{mt} and \texttt{pe} it is possible to reorder them arbitrarily, as explained in the previous section. In Figure~\ref{fig:orders_plots} we visualize three different kinds of orderings: the one produced by human post-editors, a random ordering, and the ordering obtained by processing the sentence left-to-right. 

We show, in the vertical axis of Figure~\ref{fig:orders_plots}, in which position of the sentence an action is applied, relatively to the other actions of the same sample. On the left-hand plot we display two samples. Sample A contains 3 actions, applying the leftmost action followed by the rightmost and finally an action applied in a sentence position somewhere in between the first two. This could be generated from random shuffling or human-order, but never from the artificial left-to-right order. On the other hand, Sample B contains 5 actions which could have been ordered by any of the three methods. In practice, 2.0\% of the human-ordered samples follow sample A, and 4.2\% follow sample B.

On the right-hand plot we visualize the average line for each of the three orders. We can see that human actions tend to follow a left-to-right order, but not as strictly as the artificial left-to-right. Shuffled sequences are equally likely to start on the far left or far right of the sentence.

Relative orderings displayed in Figure~\ref{fig:orders_plots} can be represented as a permutation of the sequence (\textit{0, 1, ..., \#actions}), i.e. Sample A would be (\textit{0, 2, 1}) and Sample B (\textit{0, 1, 2, 3, 4}). This way it is possible to use Kendall's $\tau$ distance \cite{kendall1938new} to measure how far we are from a pure left-to-right order. We show this in Table~\ref{tab:orderings_stats}, together with the percentage of actions which are a jump-back (applied to a position in the sentence to the left of the previous action) requiring a jump of at least 1 or 4 tokens. We confirm that the human-order is nearly left-to-right, but with some deviations.

\begin{table}[h]
\centering
\small
\begin{tabular}{lrrr}
\toprule
          & Jump-Back & JB$\geq$4 & Kendall's $\tau$ \\
\midrule
\textit{l2r}       & 0.00\%       & 0.00\%                & 0.00 \\
\textit{shuff}     & 39.43\%   & 21.34\%            & 0.48       \\
\textit{h-ord} & 14.87\%   & 4.26\%             & 0.16         \\
\bottomrule
\end{tabular}
\caption{Statistics of action orders following left-to-right, random shuffle and human-order. Jump-Back counts actions applied to any position before the previous action, whereas JB$\geq$4 requires a jump of at least 4 tokens. Kendall's $\tau$ distance is measured between the sequence (\textit{0, 1, ..., \#actions}) and its permutation matching the order of the actions in the sentence.}
\label{tab:orderings_stats}
\end{table}

\begin{figure}[h]
    \centering
    \includegraphics[width=\columnwidth]{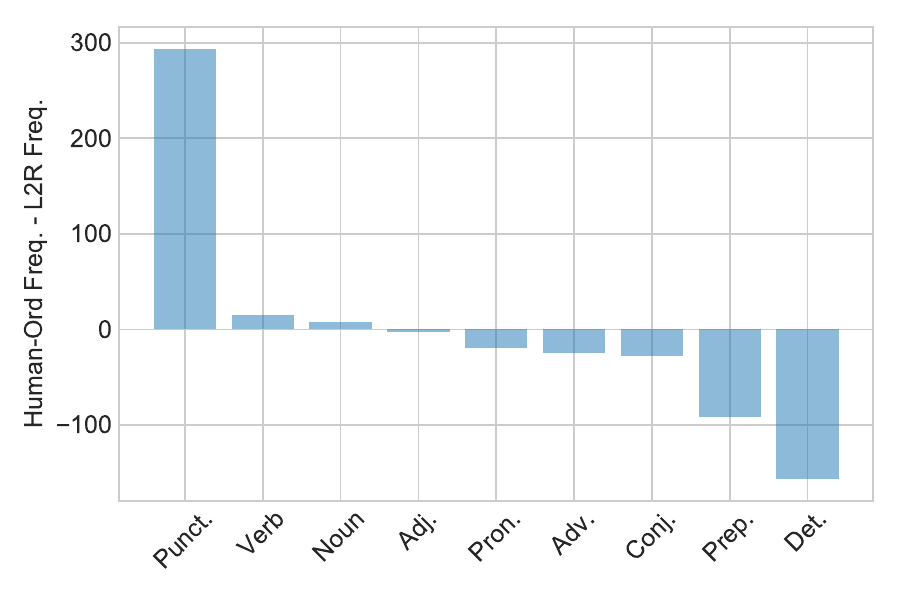}
    \caption{Part-of-speech tags preferred by humans as a first action --- positive values indicate tags preferred by humans, negative values indicate tags preferred by left-to-right order. We count occurrences of each word as first action in both human action sequences and left-to-right sequences. We subtract humans counts by left-to-right counts, discard words with a difference lower than 5, and group results by part-of-speech tag.  }
    \label{fig:pos_tags}
\end{figure}

In Figure~\ref{fig:pos_tags} we visualize the words preferred by human editors as first action. We count how many times each word is picked as the first action (without discriminating \texttt{DEL} and \texttt{INS}), both for human order and left-to-right order. We subtract human occurrences by left-to-right occurrences, keeping only words with a difference of at least 5, and group them by part-of-speech tags. We can see that humans prefer to begin with punctuation, whereas a left-to-right order favours determinants, which tend to appear in the beginning of the sentence.

\section{Model}
Inspired by recent work in non-monotonic generation \cite{stern2019insertion,gu2019insertion,emelianenko_elena_voita2019unconstrained_order}, we propose a model that receives a \hspace{0cm} \texttt{src}, \hspace{0cm} \texttt{mt} pair and outputs one action at a time. When a new action is predicted, there is no explicit memory of previous time-steps. The model can only observe the current state of the \texttt{mt}, which may have been changed by previous actions of the model.

\begin{figure}[t]
    \centering
    \includegraphics[width=\columnwidth]{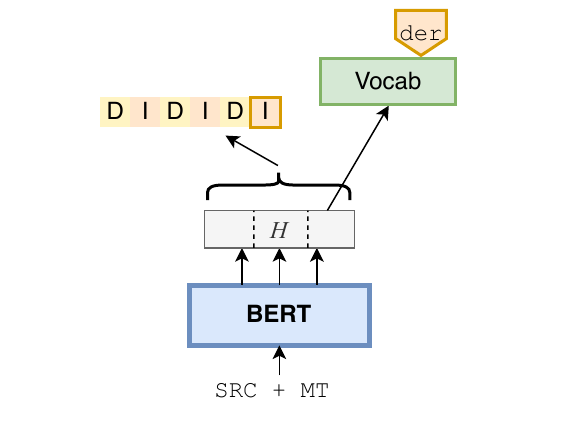}
    \caption{Our proposed model for automatic post-editing. A BERT-Encoder receives as input \texttt{src} and \texttt{mt} to produce a hidden representation $H$. We apply a linear transformation to the full $H$, obtaining a probability distribution over all possible actions. If the chosen action is an \texttt{INS}, we obtain a distribution over the vocabulary by applying another linear transformation to $H$'s row of the chosen action position.
    }
    \label{fig:model}
\end{figure}

This model is based on a Transformer-Encoder pre-trained with BERT \cite{devlin-etal-2019-bert}. After producing one hidden state for each token, a linear transformation outputs two values per token: the logit of deleting that token and of inserting a word to its right. Out of all possible \texttt{DEL} or \texttt{INS} positions, the most likely operation is selected. A special operation is reserved to represent End-of-Decoding. If an \texttt{INS} operation is chosen, we still need to choose which token to insert. Another linear transformation is applied to the hidden state of the chosen position.
We obtain a distribution over the vocabulary and select the most likely token. Figure~\ref{fig:model} illustrates this procedure.

After a \texttt{DEL} or \texttt{INS} is applied, we repeat this procedure using the updated \texttt{mt}. Decoding can end in three different ways:
\begin{itemize}
\item When the \texttt{STOP} action is predicted;
\item When the model enters a loop;
\item When a limit of 50 actions is reached.
\end{itemize}

Once decoding ends, the model outputs the final post-edited \texttt{mt}.

\paragraph{Model details.} We use BERT's implementation from \newcite{Wolf2019HuggingFacesTS} together with OpenNMT \cite{klein-etal-2017-opennmt}, both based on PyTorch \cite{paszke2019pytorch}. The pretrained BERT-Encoder contains 12 layers, embedding size and hidden size of 768.

We begin with an input sequence $x$. This sequence is the concatenation of:

\hfil \texttt{{src + [SEP] + \textless S\textgreater \hspace{0cm} + mt + \textless T\textgreater}} 

\noindent where \texttt{\textless S\textgreater} and \texttt{\textless T\textgreater} are auxiliary tokens used to allow \texttt{INS} in position 0 and to represent End-of-Decoding.  Tokens before and after \texttt{[SEP]} have a different segment embedding, to help differentiate between \texttt{src} and \texttt{mt} tokens. Let $N$ be the length of $x$ after applying a BERT pre-trained tokenizer \cite{Wolf2019HuggingFacesTS}.  This sequence is the input of the BERT-Encoder with hidden dimension $h=768$. The output is a matrix $H \in \mathbb{R}^{N \times h}$. We call each possible \texttt{DEL} position and each \texttt{INS} position an \textbf{edit-operation}. Note that this does not yet include the choice of a token from the vocabulary. For each token in the partially generated sentence, we obtain the logit of \texttt{INS} (on the position to its right) and \texttt{DEL} (of the token itself) using a learnable matrix $W \in \mathbb{R}^{h \times 2}$. The distribution probability over all possible edit-operations is defined as:

\begin{equation}
    p(\mathrm{edit\_op}) = \mathrm{softmax}(\mathrm{flatten}(HW))
\end{equation}

To represent the End-of-Decoding operation, we use the action \texttt{DEL\textless T\textgreater}. All unavailable actions are masked: \texttt{DEL\textless S\textgreater}, \texttt{INS-after\textless T\textgreater}, and edit-operations on \texttt{src}. When the model predicts an \texttt{INS}, a token is then predicted for that position. Let $i$ be the chosen position, $h_i \in \mathbb{R}^{h}$ the i\textsuperscript{th} row in $H$, and $V \in \mathbb{R}^{|v| \times h}$ the matrix mapping to all tokens in a vocabulary of size $v$:

\begin{equation}
    p(\mathrm{token} \mid \mathrm{edit\_op}) = \mathrm{softmax}(Vh_i)
\end{equation}

The predicted action is applied and we repeat this procedure using the updated $x$. Since no history of previous actions is kept, this opens the possibility of entering a loop. To handle loops, when we re-visit a state $x$ we stop decoding. 
Alternatively, we tried applying the $N$\textsuperscript{th} most likely action on the $N$\textsuperscript{th} visit to a given $x$, but this degraded performance slightly. 

\section{Training}

\begin{table*}[t]
\centering
\small
\begin{tabular}{lcccccccc}
\toprule
 &
  \multicolumn{2}{c}{dev 2016} &
  \multicolumn{2}{c}{test 2016} &
  \multicolumn{2}{c}{test 2017} &
  \multicolumn{2}{c}{test 2018} \\
      & TER $\downarrow$   & BLEU $\uparrow$ & TER $\downarrow$ & BLEU $\uparrow$ & TER $\downarrow$ & BLEU $\uparrow$ & TER $\downarrow$ & BLEU $\uparrow$ \\ \cline{2-9}
\begin{tabular}[c]{@{}l@{}}\ac{MT} baseline\\ (uncorrected)\end{tabular} &
  24.81 &
  62.92 & 
  24.76 &
  62.11 &
  24.48 &
  62.49 &
  24.24 &
  62.99 \\
  \midrule
\multirow{2}{*}{\textit{l2r}} &
 22.33 & 
 67.04 & 
 \textbf{22.53} & 
 \textbf{66.23} & 
 \textbf{22.63} & 
 \textbf{65.84} &
 22.97 & 
 65.49  \\
& \scriptsize ($\pm$0.13) & 
\scriptsize ($\pm$0.11) & 
\scriptsize \textbf{($\pm$0.26)} & 
\scriptsize \textbf{($\pm$0.26)} & 
\scriptsize \textbf{($\pm$0.3)} & 
\scriptsize \textbf{($\pm$0.29)} &
\scriptsize ($\pm$0.20) & 
\scriptsize ($\pm$0.26)  \\ [.2cm]
\multirow{2}{*}{\textit{shuff}} &
 22.47 &
 66.74 &
 22.87 &
 65.89 &
 23.24 &
 65.14 &
 22.94 &
 65.39 \\
&\scriptsize ($\pm$0.15) &
\scriptsize ($\pm$0.22) &
\scriptsize ($\pm$0.23) &
\scriptsize ($\pm$0.28) &
\scriptsize ($\pm$0.25) &
\scriptsize ($\pm$0.24) &
\scriptsize ($\pm$0.12) &
\scriptsize ($\pm$0.18) \\ [.2cm]
\multirow{2}{*}{\textit{h-ord}} &
 \textbf{22.15} &
 \textbf{67.19} & 
 22.65 & 
 66.15 & 
 22.75 & 
 65.63 & 
 \textbf{22.70} & 
     \textbf{65.72} \\
&\scriptsize \textbf{($\pm$0.23)} &
\scriptsize \textbf{($\pm$0.15)} & 
\scriptsize ($\pm$0.16) & 
\scriptsize ($\pm$0.19) & 
\scriptsize ($\pm$0.08) & 
\scriptsize ($\pm$0.04) & 
\scriptsize \textbf{($\pm$0.15)} & 
\scriptsize \textbf{($\pm$0.22)} \\
\midrule
\begin{tabular}[c]{@{}l@{}}\newcite{correia2019simple}\\ (seq2seq BERT)\end{tabular} &
  — &
  — &
  18.05 &
  72.39 &
  18.07 &
  71.90 &
  18.91 &
  70.94 \\
 \midrule
\begin{tabular}[c]{@{}l@{}}\newcite{berard-etal-2017-lig}\\ (actions)\end{tabular} &
  23.07 &
  — &
  22.89 &
  — &
  23.08 &
  65.57 &
  — &
  — \\
 \bottomrule
\end{tabular}
\caption{Results on development set and test sets used in WMT 2018 APE shared task. We show our system's performance trained by each of the three proposed orderings, and two other models for comparison. \newcite{correia2019simple} is a monotonic model following the sequence-to-sequence architecture and pre-trained on BERT (seq2seq BERT).  \newcite{berard-etal-2017-lig} predict a sequence of actions in a left-to-right order.
}
\label{tab:orders_dev_testset}
\end{table*}

During training, the model may have several correct actions to choose from, even if we only consider actions following a minimum edit distance path. We compare different ground-truth action sequences based on minimum edit actions, all arriving at the same \texttt{pe}:

\begin{itemize}
\item Left-to-right (\textit{l2r});
\item Randomly shuffled (\textit{shuff});
\item Human-ordered (\textit{h-ord}).
\end{itemize}

Minimum edit actions are generated using the dynamic programming algorithm to compute Levenshtein distance. The algorithm is set to output left-to-right actions, but since its output contains no redundant actions, they can be arbitrarily re-ordered. One simple way to re-order the actions is by randomly shuffling them. A more sophisticated alternative consists in matching each minimum-edit action to a human action, as described in \S\ref{sec:preprocess}.

We also experimented with unfiltered human actions. However this resulted in significantly inferior performance, possibly due to the hesitations made by humans typing, who may add and delete words unnecessary for the final \texttt{pe}.

\paragraph{Training details.}
We train the model by maximizing the likelihood of the action sequences provided as ground truth. Following \newcite{correia2019simple} we use Adam \cite{kingma2014adam} with a triangular schedule, increasing linearly for the first 5,000 steps until $5\times10^{-5}$, applying a linear decay afterwards. BERT components have $\ell_2$ weight decay of $10^{-4}$. We apply dropout \cite{srivastava_JMLR_dropout} with $p_{drop}=0.1$ and, for the loss of vocabulary distribution, label-smoothing with $\epsilon=0.1$ \cite{pereyra2017regularizing}. We use batch size of 512 tokens and save checkpoints every 10,000 steps.

\section{Experiments}

We compare the effect of using different action orders on the development set and test sets of WMT 2018 \ac{APE} shared task \cite{chatterjee2018findings}.

By using only training data overlapping with WMT's training sets (as described in \S\ref{sec:data_subsection}), we are able to use WMT's development and test sets for evaluation. This allows to compare the performance of our model with that of previous submissions. Note however that our systems are in disadvantage, due to being trained on fewer data: out of the original 23,000 training samples we only found 16,068 in \newcite{specia2017translation}.

\subsection{Performance of models}

We explore  three different ways to order the actions provided by minimum edit distance: \textit{l2r}, \textit{shuff} and \textit{h-ord}. For each run, we pick the best checkpoint measured by TER in the development set, and evaluate on 3 test sets. Table~\ref{tab:orders_dev_testset} shows the average and standard deviation of 5 runs. Depending on the dataset chosen, the best performance is achieved by either \textit{l2r} or \textit{h-ord}, with small variations between the two. Random shuffling is consistently worse than the alternatives, by a margin of around 0.3 TER. All three alternatives significantly improve the uncorrected \ac{MT} baseline.

To compare with existing results, we choose two models. \newcite{correia2019simple} use an architecture based on a monotonic autoregressive decoder (factorized in a left-to-right order). They propose a strategy that leverages on the pretrained BERT transformer \cite{devlin-etal-2019-bert}, achieving performance gains with it. Monotonic autoregressive models typically achieve a superior performance than their non-monotonic and non-autoregressive counterparts \cite{Zhou2020Understanding}.
\newcite{berard-etal-2017-lig} propose a model that predicts a sequence of actions, which is closer to our approach, although they impose a left-to-right order. As expected, we do not outperform the results of the monotonic autoregressive model. However, we beat B{\'e}rard's action-based model, even though we use a smaller training set due to requiring keystrokes (16,086 samples instead of 23,000). This gain could be due to the pretraining of the BERT encoder used in our model, but also because of a largely different architeture (e.g. we use a Transformer encoder instead of a LSTM).

\subsection{Learned orderings}

\begin{figure}[t]
    \centering
    \includegraphics[width=\columnwidth]{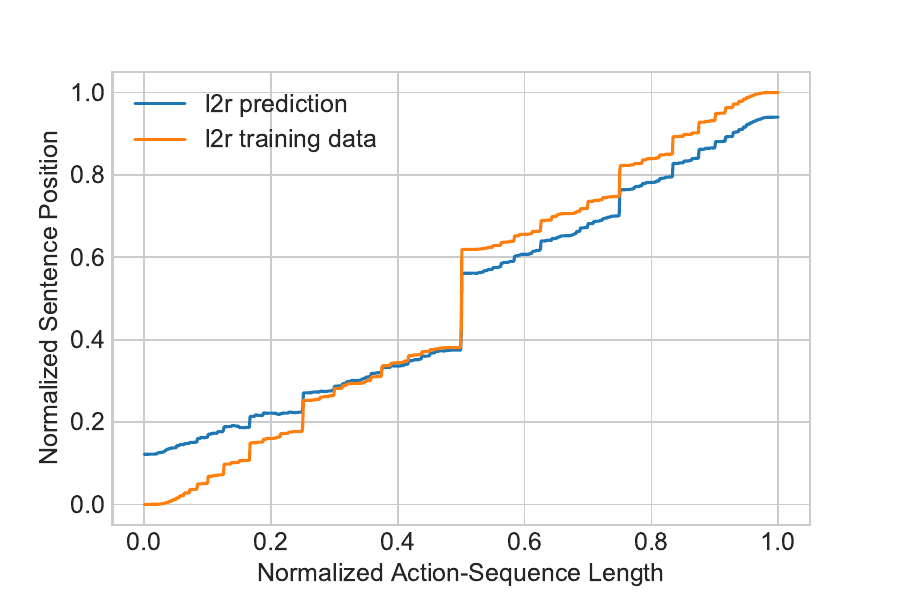}
    \caption{Relative positions of actions in a sequence, as explained in Figure~\ref{fig:orders_plots}. Comparing the original full left-to-right training data curve (orange) with the model predictions (blue), we see that the model became slightly non-monotonic.}
    \label{fig:l2r_pred}
\end{figure}

\begin{table}[!b]
\small
\begin{tabular}{lrrrr}
\toprule
      & Kendall's $\tau$ & $\Delta$ K's $\tau$   & \%loops & \%do-noth  \\
\midrule
\textit{l2r} & 0.04  &   +0.04    & 15.6    & 10.5      \\
\textit{shuff} & 0.50  &    +0.02   & 12.9    & 11.2       \\
\textit{h-ord}   & 0.16 &   0.00    & 16.0    & 9.8         \\
\bottomrule
\end{tabular}
\caption{Statistics on the actions predicted by the 3 different models, measured on a single run of the development set. $\Delta$~K's~$\tau$ refers to the difference between Kendall's $\tau$ distance of the model's output and of the corresponding training data. \%loops counts samples that entered a loop, and \%do-noth counts samples where the first predicted action was STOP.}
\label{tab:output_stats}
\end{table}

Regarding the orderings learned by our model, they largely resemble the behaviour of the training data. Similar values for Kendall's $\tau$ distance indicate a similar amount of non-monotonicity in each of the three scenarios, as seen in Table~\ref{tab:output_stats}. The only exception is the left-to-right model which, unlike the training data, becomes slightly non-monotonic during inference time. This is shown by the increase in Kendall's $\tau$ distance and illustrated by Figure~\ref{fig:l2r_pred}. This imprecision in the decoding order may be expected since the model does not have an explicit memory of what has already been done.

\section{Related Work}

\paragraph{Non-monotonic models.}
Recent work explored alternatives for neural sequence generation that do not impose a left-to-right generation order. On the one hand, this allows for bidirectional attention to both left and right context of the token being generated. On the other hand, it is a more challenging task since it implies learning a generation order from a number of possibilities that grows exponentially. Generation order is usually treated as a latent variable, and our work differs in that we use supervision from human post-editors.

\newcite{gu2019insertion} propose an insertion-based model which avoids re-encoding by using relative attention, and has two ways of learning order: one using pre-defined orders, the other searching for orders that maximize the sequence likelihood, given the current model parameters. \newcite{emelianenko_elena_voita2019unconstrained_order} train using sampled orders instead,
to better escape local optima. They also drop the relative attention mechanism together with its better theoretical bound on time complexity -- showing that, in practice, inference remains feasible.

\newcite{welleck2019non} propose a model that generates text as a binary tree. They learn order from a uniform distribution that slowly shifts to search among the model's own preferences, or alternatively using a deterministic left-to-right oracle.

\newcite{lawrence-etal-2019-attending} use placeholders to represent yet-to-insert tokens, allowing for bidirectional attention without exposing future tokens. Decoding is either done left-to-right or by picking the most certain prediction. Alternatively all tokens can be decided in parallel, but with significant loss in performance.

\paragraph{Non-autoregressive models.} Another class of models focuses on parallel decoding of multiple tokens, moving away from the traditional autoregressive paradigm. This unlocks faster inference, but brings the difficult challenge of learning dependencies between tokens \cite{gu2018nonautoregressive}. \newcite{stern2019insertion} explore both non-monotonic autoregressive and non-autoregressive decoding with the Insertion Transformer. They use loss functions that promote either left-to-right order, a uniform distribution or a balanced binary tree for maximal parallelization.

The recently proposed Levenshtein Transformer \cite{gu2019levenshtein} introduces a Delete operation, and can generate or refine text by iterating between parallel insertions and parallel deletions --- allowing to tackle the task of \ac{MT} and also \ac{APE}. \newcite{ruis2020insertion} add a Delete operation to the Insertion Transformer and evaluate on artificial tasks. Our work differs in that we keep our model autoregressive, tackle the non-monotonicity by providing supervision to the order, analyze learned orders and focus on the \ac{APE} task.

In general, this class of models is difficult to train and relies on several tricks. Knowledge distillation can bring improvements \cite{Zhou2020Understanding}, recently allowing Levenshtein Transformer to close the gap in translation quality between autoregressive monotonic and non-autoregressive models. In our setup, the tools proposed by \newcite{Zhou2020Understanding} to measure data complexity could be used, for instance, for filtering out samples which are too complex.

\paragraph{Automatic post-editing.} \ac{APE} was initially proposed to combine rule-based translation systems with statistical phrase-based post-editing \cite{simard-etal-2007-post-editing}. As the quality of MT systems improves, there is less benefit in post-editing its mistakes, in particular if the MT system is trained on in-domain data. Current neural MT systems tend to generate very fluent output, therefore to fix their mistakes it is not enough to  look at the \texttt{mt} output, but more deeply seek information from the \texttt{src} to fix adequacy errors. Top-performing systems for post-editing currently rely on tricks such as round-trip translation \cite{junczys2016log-ape} to increase dataset size, leveraging on pre-trained models \cite{correia2019simple} and using conservativeness penalties \cite{lopes2019unbabel} to avoid over-editing. \newcite{berard-etal-2017-lig} post-edit by predicting a sequence of actions with an imposed left-to-right order. Another recent work directly models edits, without including order information but allowing to re-use edits in unseen contexts \cite{yin2018learning}.

\paragraph{Human post-editing.} Previous work has explored keystrokes to understand the behavior of human editors. \newcite{OBrien2006} investigates the relationship between pauses and cognitive effort, while later research \cite{LaCruz2012,LaCruz2014} examines keystroke logs for the same effect. \newcite{specia2017translation} introduce a dataset of human post-edits, containing information on keytrokes. Recently it was shown that detailed information from post-editing, such as sequences of edit-operations combined with mouseclicks and waiting times, contain structured information \cite{gois2019translator2vec}. The same work provides evidence that this kind of information allows to identify and profile editors, and may be helpful in downstream tasks.

\section{Conclusions}

In this work we explored different ways to order the edit operations necessary to fix mistakes in a translated sentence. In particular, we studied which orderings are produced by humans, and whether they can be used to guide the training of a non-monotonic post-editing system.

We found that humans tend to use nearly left-to-right order, although with exceptions, such as preferring to fix punctuation and verbs first. We then proposed a Transformer-based model pre-trained with BERT that learns to automatically post-edit translations in a flexible order. We learned this model in three different ways: by supervising it with orderings performed by humans, by using a left-to-right ordering, or with random  orderings. In all three settings, the model outperformed the uncorrected machine translation baseline and a previous system also designed to predict actions \cite{berard-etal-2017-lig}. 

Training the model with human orderings achieved performance equivalent to left-to-right, or even superior. The random order consistently yielded slightly lower results. The model learned to mimic the proposed orders in all three cases.

\section*{Acknowledgments}

We  would  like to thank Iacer Calixto, Ant\'onio Lopes, F\'abio Kepler, Sean Welleck, Nikita Nangia, and the anonymous reviewers for their insightful comments. 
This work was partially supported by the EU/FEDER programme under PT2020 (contracts 042671 and 038510)  
and by  the  European  Research  Council  (ERC  StG  DeepSPIN  758969).

\acrodef{MT}{machine translation}
\acrodef{APE}{automatic post-editing}
\acrodef{IT}{Information Technology}

\bibliography{eamt20}
\bibliographystyle{eamt20}

\end{document}